\documentclass[a4paper, 10pt, conference]{ieeeconf}      

\IEEEoverridecommandlockouts                              
                                                          
\usepackage{amsmath}
\usepackage{amssymb}  
\usepackage{graphicx} 

\def\etal{\mbox{et al.}}

\def \RR {{\mathbb R}}

\begin{document}

\title{Positioning aiding using {LiDAR} in {GPS} signal loss scenarios*}

\author{Szymon Krupi\'nski$^{1}$, Francesco Maurelli$^{1}$\\ 
${}^{1}$Jacobs University Bremen, Germany\\
\{s.krupinski,f.maurelli\}@jacobs-university.de\\
\thanks{*The research leading to these results has received funding from the European Union Horizon2020 Programme - Marie Sk\l{}odowska-Curie Action - under grant agreement No. 709136 TIC-AUV}
}
\maketitle

\begin{abstract}

In the presented scenario, an autonomous surface vehicle (ASV) equipped with a laser scanner navigates on a inland pathway surrounded and crossed by man-made structures such as bridges and locks. {GPS} receiver present on board experiences signal loss and multipath reflections in situation when the view of the sky is obscured by a bridge or tall buildings. In both cases, a potentially dangerous situation is provoked as the robot has no or inaccurate positioning data. A sensor data processing scheme is proposed where these gaps are smoothly filled in by positioning data generated from scan matching and registration of the laser data. This article shows preliminary results of positioning data improvement during trials in harbor-river environment.

\bigskip

\end{abstract}

\section{Introduction} \label{sect:intro}

\subsection{Motivation}


In the domain of underwater robotics, the standard suite of navigational sensors include an inertial measurement system, Doppler velocity log and a {GPS} receiver for initial surface positioning fixes. It is sometimes augmented with acoustic positioning beacons such as long- or ultra short base line ({LBL}/{USBL}) which provides low frequency and relatively noisy estimate but is free from drift accumulation over time. Subject to the quality of the chosen components, this setup tends to provide reasonable positioning accuracy in zones covered by acoustic localization. In its absence, positioning quality quickly deteriorates during the mission.

For the surface vehicles, the localization is greatly simplified due to the near constant availability of the {GPS} signal. Its ubiquity causes that the alternative solution is rarely sought. Contrary to the autonomous underwater vehicle {AUVs}, surface vehicle very rarely carry a DVL and might not be equipped even with a simpler mechanical equivalent. This restricts what can be done in case of {GPS} signal failure even further. And this failure is a fairly regular occurrence in man-made (harbor) or complex natural environments (canyon, mangrove). While today autonomous operations in harbours, rivers and navigational canals represent only a small fraction of work done, it is precisely in these environments where their unreliability can lead to costly and dangerous consequences, such as a collision.

The proposition in this article is to turn to the payload sensor for additional positioning clues. As the illustrating example, a data-rich 3-D laser scanner is used. In case of {AUVs}, the same idea could be applied to a bathymetric or imaging sonar.

The ultimate motivation of developing a {LiDAR}-equipped vehicle with robust navigation is to use it for autonomous data gathering in harbor, river and coastal environments. In the future, this technology can serve to develop autonomous water taxis, ferries and other ships.

\subsection{Related work}

A number of well established methods exist for positioning without constant availability of {GPS} data. Underwater vehicle positioning is a notable application, since {GPS}, or, for that matter, any other signal using electromagnetic ({EM}) waves, cannot penetrate water beyond a limited skin depth. Thus, the initial {GPS} fix is taken on the surface and then, the positioning is carried out using a combination of dead reckoning, integration of inertial measurements and acoustic distance fixes, if available.

A couple of filtering techniques have achieved the status of a near standard positioning solution in marine robotics and similar domains, namely Kalman (Extended or Unscented) and Particle filtering \cite{maurelli2009sonar}, \cite{mallios2010ekf}. They permit integrating several heterogeneous sensor measurements of complimentary characteristics and calculating the optimal estimation of the current vehicle state.

Several advanced, mostly experimental autonomous platforms are also dotted in the complimentary environment mapping algorithms, together constituting a simultaneous localization and mapping {SLAM}. The technique was also applied to underwater robots by Ribas~\cite{ribas2008underwater}, Mallios~\cite{mallios2010ekf} and others. While SLAM represents a complete localization solution, it creates a significant memory overhead due to the need to maintain a map of the environment and the computation needs due to constant new sensor data and map updates. In case of survey vehicles, creating a product-grade map of the scene for immediate navigation purpose would be impractical. Thus, some authors propose a localisation on an (partial) existing map, notably using techniques of scan matching \cite{maurelli2009sonar}. Matching a current scan to a global map yields a global position candidate. The approach taken in this work makes use of relative matching of consecutive scans without a global map.

{LiDAR}, or 3-D laser scanners provide a measurement of distance along an array of laser beams typically mounted on a rotating head, so that they can sweep a large volume of the environment. They are commonly used in autonomous driving \cite{maurelli20093d} but they are also making their way underwater where they enable such applications as autonomous or milimetric-precision survey \cite{mcleod2013autonomous}. 3-D scanners produce a considerable volume of data in form of structures clouds of 3-D points, although some devices can also output depth images. In addition to the geometric information, the intensity of laser reflection is often recorded for every point.

In addition to sonar and LiDAR, vision-based ASV navigation has been explored, for example by Dunbabin~\etal \cite{Dunbabin2008VisionbasedDU}, Wang~\etal~\cite{Wang2015VisionBasedTC}, Heidarsson and Sukhatme~\cite{Heidarsson2011ObstacleDF}. Despite producing information-rich data at a high frequency, standard imaging techniques do not provide the exact geometric information about the environment. Thus, with the decreasing sensor prices, LiDARs become more commonplace in autonomous vehicles, including ASVs.

The idea of calculating relative displacement from the incremental {LiDAR} scan matching is not new. Tang~\etal \cite{tang2015lidar} makes use of such technique for indoor navigation of a terrestrial robot. However, data collected in a building environment has many easy to exploit characteristics, such as near-perfect planes created by the walls, short sensing distances and the existence of the ground plane. The techniques using bathymetric sonar readings and a known depth map, commonly referred to as ``terrain-based navigation'' belong to the same family, although with a slightly different geometry of the problem (as explored by Lucido~\etal~\cite{lucido1998segmentation}, Li~\etal~\cite{li2018underwater} and others).

\begin{figure}[htp]
  \centering
  \includegraphics[width=0.8\linewidth]{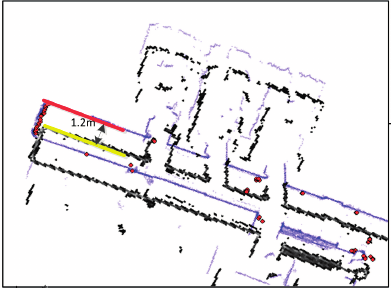}
  \caption{An example from \cite{tang2015lidar} of matching wall contours detected by a {LiDAR} device in an indoor application.}
  \label{fig:tang_lidar}
\end{figure}

\subsection{Investigated scenario}

In the presented scenario, an autonomous surface vehicle ({ASV}) navigates on a inland pathway surrounded and crossed by man-made structures such as bridges and locks. GPS receiver present on board experiences signal loss in situation when the view of the sky is obscured by a bridge. In other situations, for example navigating close to a tall building or a canal wall, the positioning signal is known to be distorted by multipath reflections, giving false position readings. In both cases, a potentially dangerous situation is provoked as the robot has no or inaccurate positioning data.

Horizontal Dilution of Precision (HDOP) gives estimate of the of the current precision due to satellite position and conditions. It can serve as a criterium of whether using {GPS} input is safe. In practice, a low HDOP does not guarantee a correct position fix. During the tests, the two positioning methods are run in parallel in order to better analyse such situation and to enable to design a future robust switching strategy.

\begin{figure}[thpb]
    \centering
    \includegraphics[width=0.49\linewidth]{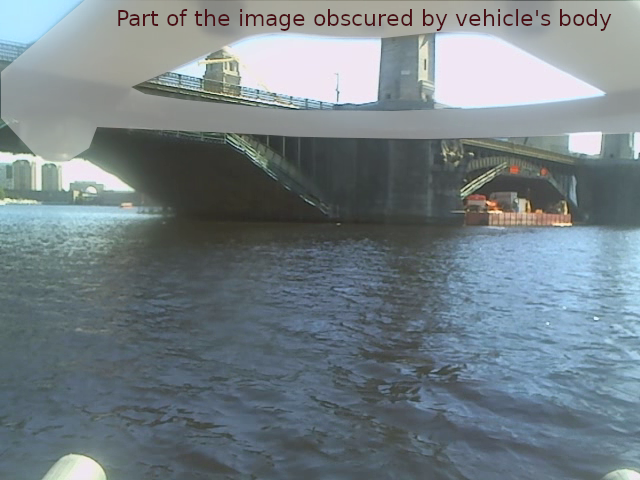}
    \includegraphics[width=0.49\linewidth]{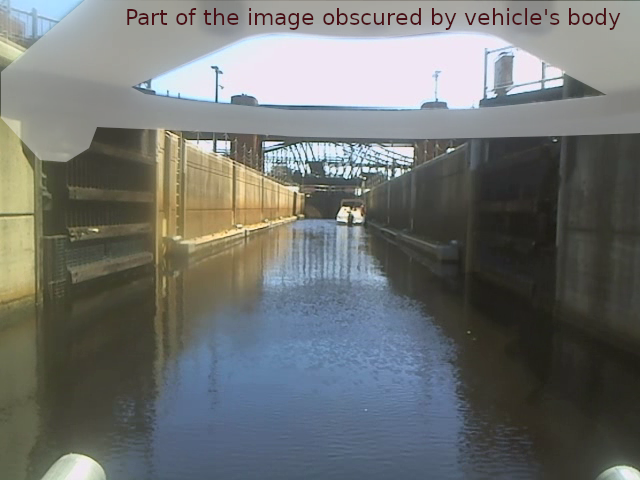}
    \caption{Still image frames from the video camera installed on board of the vehicle showing environment likely to cause GPS signal disturbance.}
    \label{fig:vidframes}
\end{figure}

\begin{figure}[htp]
  \centering
  \includegraphics[width=\linewidth]{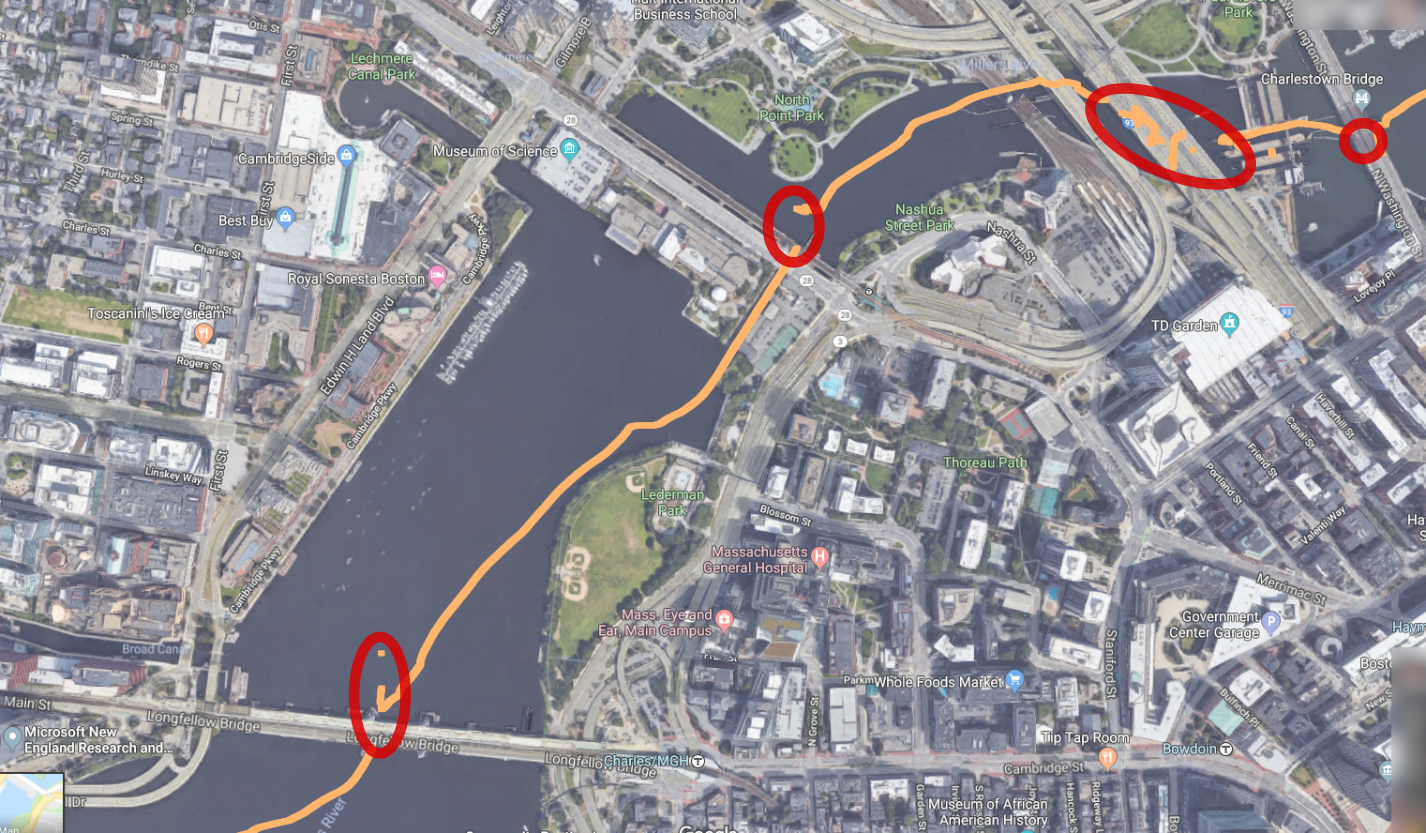}
  \caption{Red circle show the discontinuities and false indication of the {GPS} sensor as the vehicle crosses bridges and locks on the Charles river.}
  \label{fig:discerrors}
\end{figure}

\begin{figure}[htp]
  \centering
  \includegraphics[width=\linewidth]{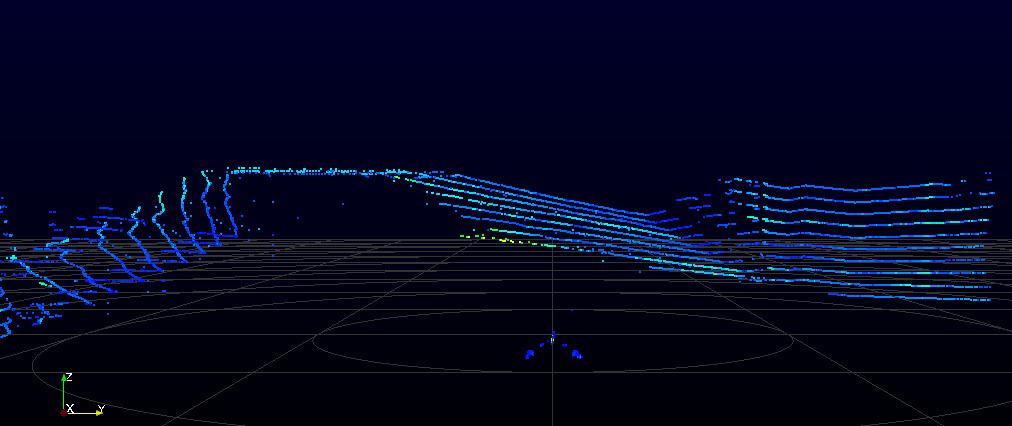}
  \caption{Part of the point cloud representing one of the bridge pillars in front of the vehicle, as seen in the left-hand-side of Fig. \ref{fig:vidframes}.}
  \label{fig:veloviewptcloud}
\end{figure}

Since the blackout of the {GPS} sensor means that there is no alternative positioning data, the technique can be validated using a parts of the vehicle's route where the {GPS} signal was of good quality and can thus serve as the ground truth.

\section{PROPOSED NAVIGATION SCHEME}


The positioning aiding discussed here is supposed to complement a generally reliable {GPS} indications. Due to the nature of the incremental scan matching, if one starts collecting scans only after a detected signal loss, the calculated position may be based on the already false last estimation. Additionally, given the difficulty to automatically detect failures using simple metrics such as {HDOP}, it is reasonable to assume that the point cloud processing has to run continuously. Given that the vehicle runs on a limited energy source, it is imperative to propose an efficient algorithm.

In general, the problem of scan matching considered here is an optimisation problem of finding a transformation (here, represented by an affine transformation $R \in \RR^{4x4}$ including rotation and translation) that minimises distances between all corresponding points $p^{S_a}_n \in  \RR^4$  and $p^{S_b}_m \in \RR^4$ (homogeneous coordinates) of two point clouds $S_a$ and $S_b$ respectively:

\begin{align*}
R^* &= \underset{R}{\operatorname{argmin}} {\sum_{\text{all pairs }(m,n)} R p^{S_a}_m - p^{S_b}_n}.
\end{align*}

\begin{figure}[htp]
  \centering
  \includegraphics[width=\linewidth]{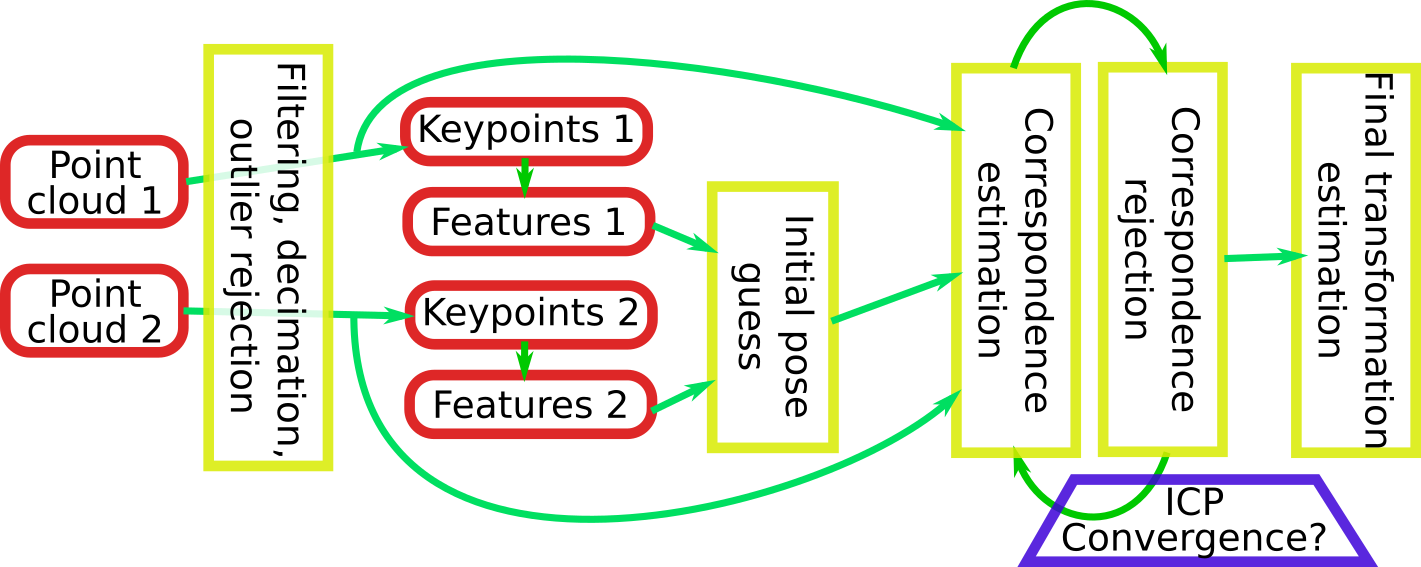}
  \caption{Method 1: Canonical {PCL} point cloud registration pipeline.}
  \label{fig:registpipe}
\end{figure}

\begin{figure}[htp]
  \centering
  \includegraphics[width=\linewidth]{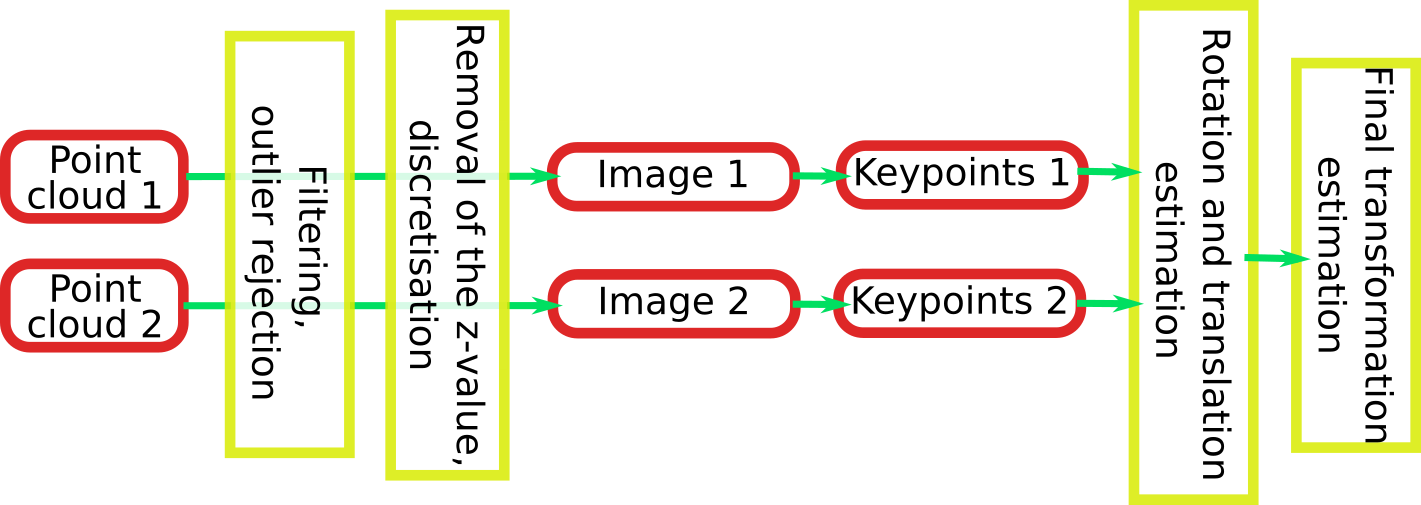}
  \caption{Method 2: Simplified, 3-D registration of scans.}
  \label{fig:custompipe}
\end{figure}

If point clouds/scans $S_a$ and $S_b$ are expressed in a frame of reference $\mathcal{A}$ and $\mathcal{B}$ respectively, then the final calculated transformation ($R^*: \mathcal{A} \rightarrow \mathcal{B}$) can then be directly used as the relative rotation/displacement between the vehicle's poses at which the scans were taken. Given a large operating scan speed (less than 0.1s to complete a $360^\circ$ scan), the deformation of the point cloud due to vehicle's motion can be neglected, contrary to the equivalent use of a rotating head sonar in \cite{ribas2008underwater}. 
For the trial campaign, two algorithms were considered. They are summarised in the table below and illustrated in Figs \ref{fig:registpipe} and \ref{fig:custompipe}.

\begin{table}[hbt]
{\begin{tabular}{p{0.47\linewidth} p{0.47\linewidth}}
   \textbf{Full 6-D registration} & \textbf{Reducted 3-D matching} \\
   Feed two point cloud captured during separate scans into the regular {PCL} registration pipeline, output: 3 components of rotation (yaw, pitch, roll) and a  translations vector & Down-project the point clouds on the x-y plane and use image-based registration to obtain 3 components of the relative motion: yaw, x- and y-translation \\
  \end{tabular}}
\label{table:refframes}
\end{table}

While the full 6-D registration permits to exploit state-of-the-art algorithms bundled with the {PCL}, it's performance is directly related to the number of points in the point cloud. In a calm weather, where roll and pitch are bounded to several degrees, trying to calculate them is often counterproductive, especially for a sensor mounted on a Wave Adaptive Modular Vessel (WAM-V) platform, which has inherent stability due to the mechanical design (www.wam-v.com/tech). It can be noted that the z-translation of the point cloud can be expected to be null. In order to speed up the calculations, the initial pose guess can be based on these criteria.

\begin{figure}[htp]
  \centering
  \includegraphics[width=0.58\linewidth]{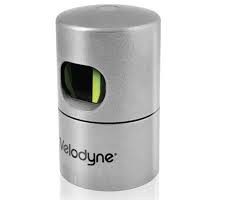}
  \caption{Velodyne 32E {LiDAR} sensor, as used during the trials, is characterised by small size.}
  \label{fig:velo32e}
\end{figure}

{PCL}'s canonical point cloud registration includes a stage of filtering, correspondence rejection and pre-aligning, before the final fine alignment is carried out using Iterative Closest-Point algorithm (ICP) \cite{rusinkiewicz2001efficient}. However, if scans from a very short period are used as input, the displacement/rotation between them is typically very low, thus they can be considered pre-aligned. For the implementation of the pre-alignment, Fast Point Feature Histograms ({FPFH}) descriptors computing algorithm \cite{RusuDoctoralDissertation} bundled with {PCL} was chosen after trying a number of alternatives.

While the initial version of the algorithm was conform to the scheme given in Fig. \ref{fig:registpipe}, some variations were tried in the field.

A simpler method was also introduced which projects all points in the cloud onto a x-y plane to form an image. The size and pixel dimensions of this image are important parameters: some points further away will be dropped if they do not fit the canvas. The resolution of the image will essentially determine the expected precision. The desired side-effect of this operation is that if there is a dense point cloud segment, e.g. from a nearby wall or bridge pillar, it will form a bright line or zone in the image, at the same time reducing significantly the number of points to process. The consecutive scan is likely to produce a similar characteristic shape; the relative rotation and translation can thus be computed with high confidence thanks to this correspondence. The resulting images resemble strongly imaging sonar captures of a sea bottom and structures.

\begin{figure}[thpb]
    \centering
    \includegraphics[width=0.49\linewidth]{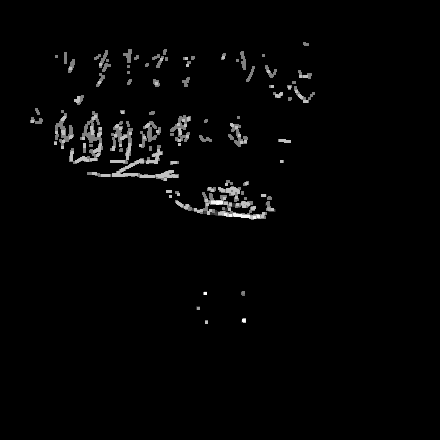}
    \includegraphics[width=0.49\linewidth]{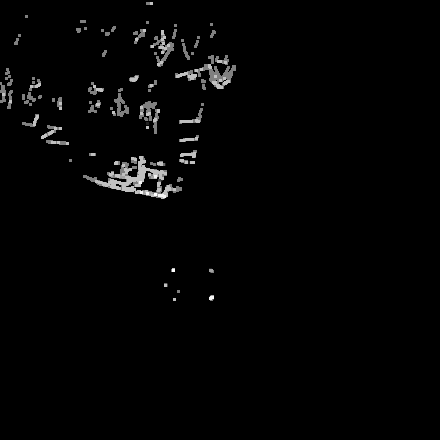}
    \caption{Results of reduction of point clouds from the dataset to 2-D images in the simple algorithm (points dilated for better visibility).}
    \label{fig:framessimpalg}
\end{figure}

\section{IMPLEMENTATION AND RESULTS}

\begin{figure}[htp]
  \centering
  \includegraphics[width=\linewidth]{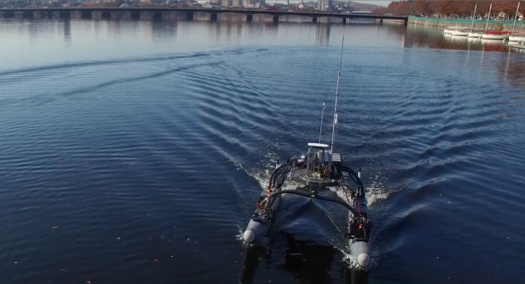}
  \caption{{REx} IV {ASV} used for data collection. During the trials, it was remotely controlled from a chase boat. Image source \cite{MITSGwebsite}}
  \label{fig:tang_lidar}
\end{figure}

The navigation algorithm was tested on a Reef Explorer IV (REx IV) vehicle belonging to the MIT Sea Grant research group navigating on the section of Charles River between Boston University Bridge and Charlestown Bridge, featuring tall city architecture, multiple bridges of different types, locks and canal walls. The vehicle (more info available at \cite{MITSGwebsite}) is of WAM-V design and carries a top mounted Velodyne HDL-32E Lidar Sensor, a camera and a {GPS} receiver. During normal operation in open space of the river, the {GPS} provides a satisfactory estimation of position, with uncertainty restricted to less than meter.

The data processing and positioning pipeline was implemented in the environment of Robotic Operating System (ROS) \cite{quigley2009ros} using Point Cloud Library (PCL, pointclouds.org) for all operations on 3-D point clouds. This library contains an extensive and slowly maturing framework for registration of point clouds which contains necessary functions to build the discussed positioning algorithms \cite{holz2015registration}.

\begin{figure}[htp]
  \centering
  \includegraphics[width=\linewidth]{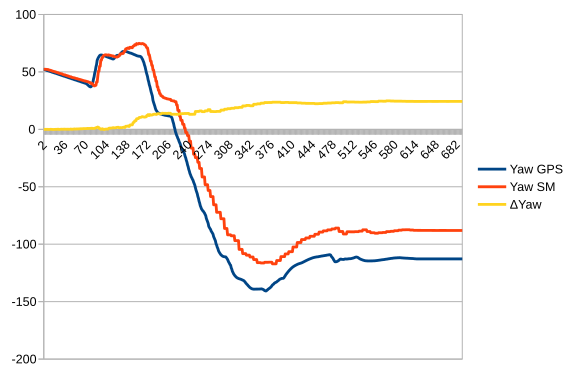}
  \caption{Yaw calculated using scan matching seems to reflect all the variations of the compass/{GPS} cap but there is a growing drift over the test period, reaching $20^\circ$ at the end.}
  \label{fig:yawanalysis}
\end{figure}

\begin{figure}[htp]
  \centering
  \includegraphics[width=\linewidth]{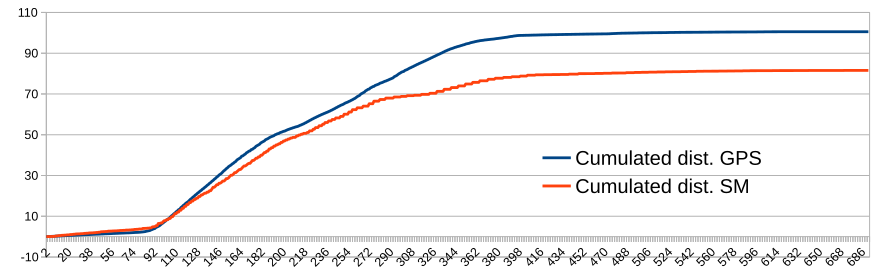}
  \caption{Travelled distance estimated by scan matching.}
  \label{fig:distanceanalysis}
\end{figure}

\begin{figure}[htp]
  \centering
  \includegraphics[width=\linewidth]{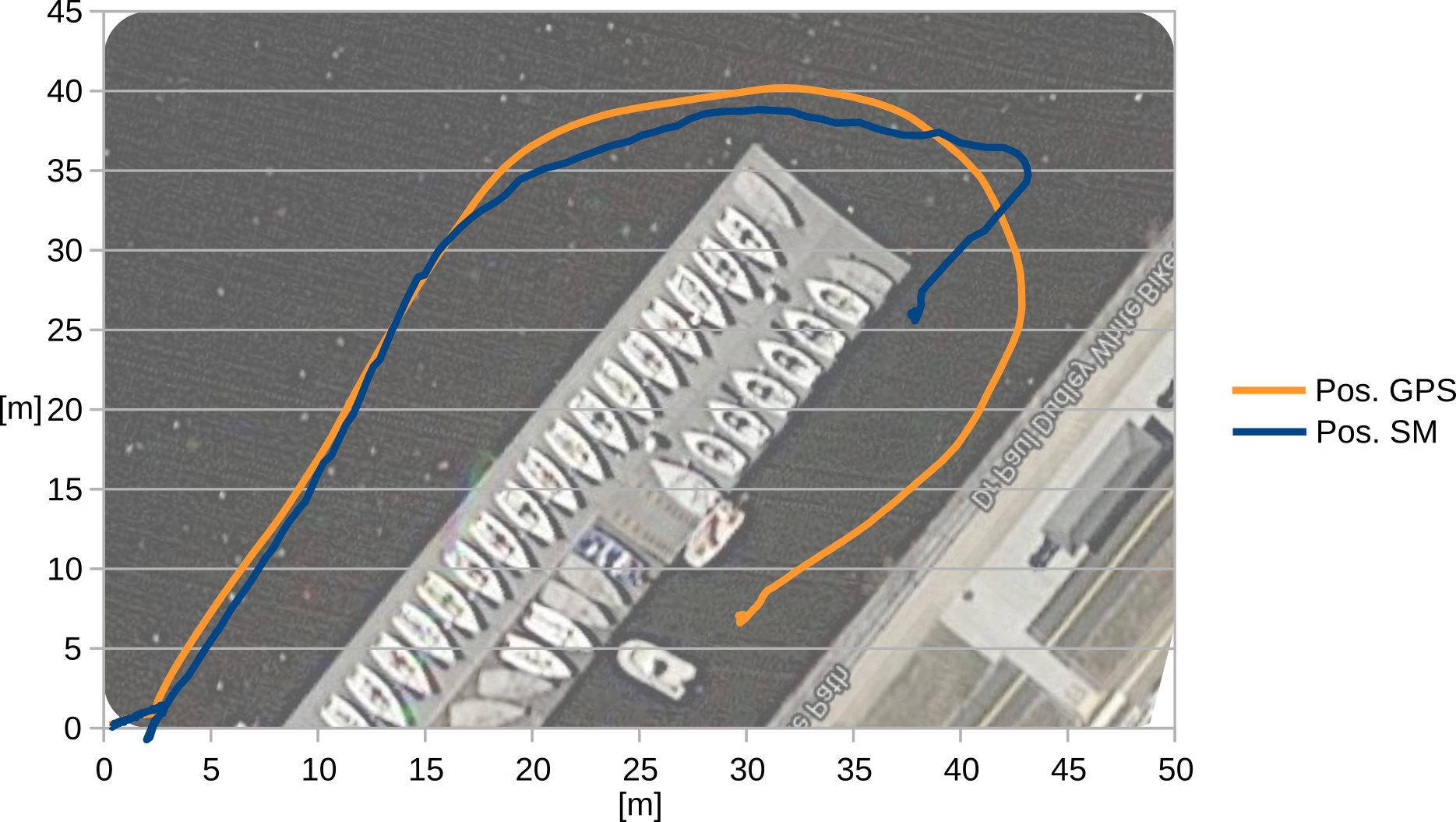}
  \caption{The experiment in which the {GPS} localisation was used as ground truth shows some imperfection in the estimation but also the potential of the scan matching technique.}
  \label{fig:dockingmap}
\end{figure}

\addtolength{\textheight}{0.0cm}   

Analysis of several collected data sets reveals some properties of the data coming from the LiDAR device:
\begin{enumerate}  
\item Almost no points are registered on the water surface
\item The clouds are relatively scarce
\item Complex structures with girders, pipes, etc. return point clouds where normal estimation is unreliable
\item HDOP is not an infallible indicator of {GPS} errors, especially in the multipath scenario
\end{enumerate}

The above observations have direct consequences for the choice and performance of the scan matching algorithms. Property 1) excludes existing algorithms which rely on the detection of the base plane. 2) and 3) render algorithms which rely on surface normals less robust, yet {FPFH}, which uses the information about normals was still the best-performing descriptor. Point 4) is a signal against using {HDOP} to select which source of localisation: {GPS} or scan-matching is to be trusted more. More work needs to be dedicated to finding an estimator which will allow detecting early stage of signal blackout.

\begin{figure}[htp]
  \centering
  \includegraphics[width=\linewidth]{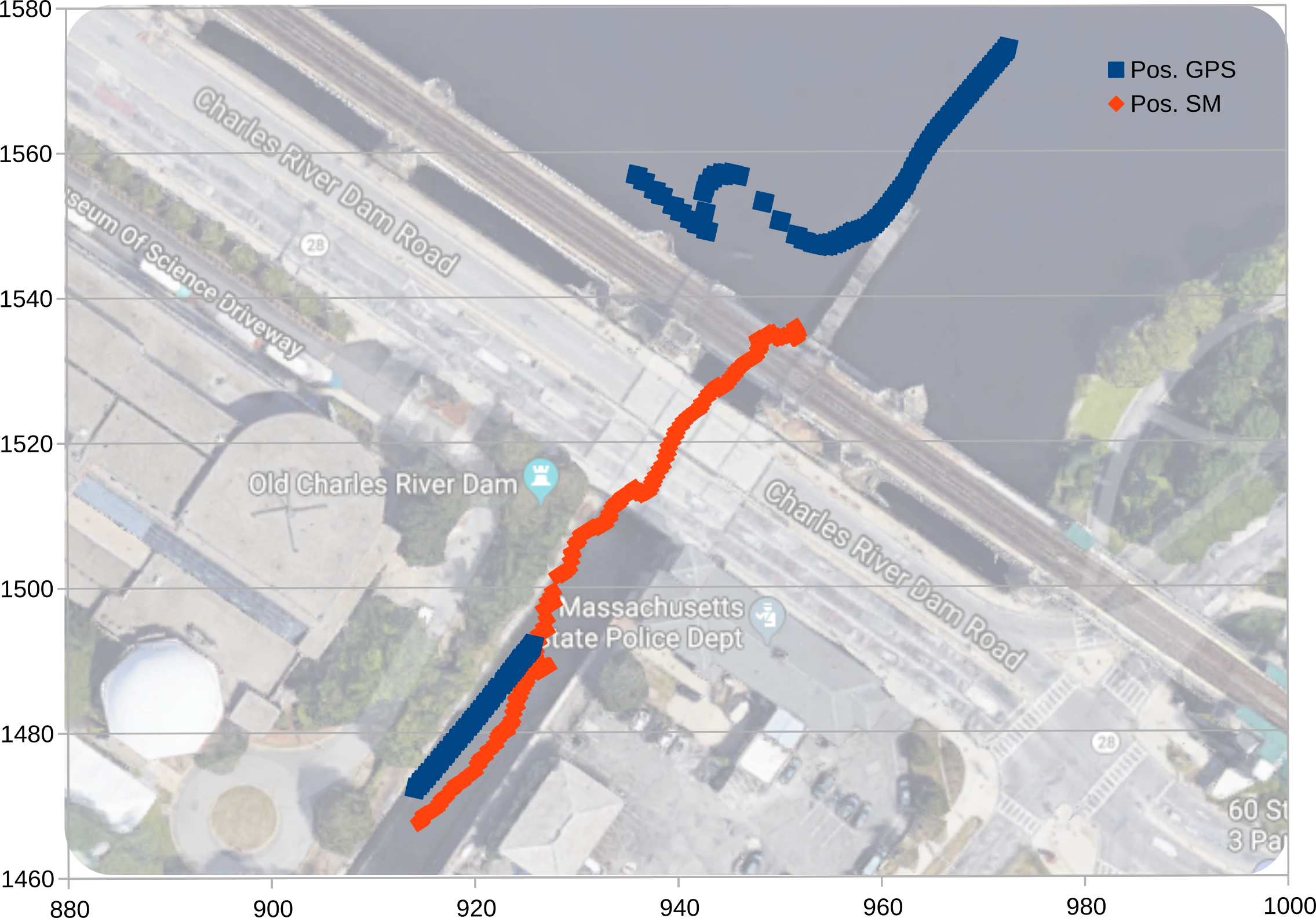}
  \caption{Crossing Charles River Dam Road caused a {GPS} blackout followed by a period of incorrect estimation smoothed through the device's internal filter. The scan matching-based positioning kept track of the ASV's progression under the bridge, albeit the reconstruction of the trajectory is rather noisy. In this trial, the 1st method was used in post-processing.}
  \label{fig:dockingmap}
\end{figure}

The performance obtained on a regular laptop {PC} used for the processing were 10 Hz and above with the simplified method and between 8 Hz - 1 Hz with full PCL processing chain. The last result varied with the number of points present in either of the scan point clouds. Depending on presence of shore or structures near the vehicle, the successful matching was achieved for 6,000 - 27,000 points per scan, with mostly failures below this range. On average, below 10\% of the points were filtered out at the outlier rejection stage. Given that the scan rate was 10Hz, the simplified method was performing in real time but its results were characterised by a lower degree of precision and higher percentage of unresolved scan matching. In good conditions, the ICP stage was virtually unnecessary, as the pre-alignement with {FPFH} produced a nearly ideal estimation. It is, however, not yet quantified what impact on the final result would eliminating ICP have.

An additional stage of processing was introduced due to the experience gained during the trials: the scans were compared w.r.t. the number of points they contained. Too big discrepancy would normally signify that the incoming scan was anomalous (e.g. due to a momentary blinding of the Velodyne sensor) and had to be dropped.

\section{CONCLUSIONS}  \label{sect:concl}

\subsection{Achieved objectives}

Despite a low degree of precision, the preliminary results show that the method can be used to render navigation in cluttered environments more robust. At this time it is difficult to achieve precision and real-time processing at the same time but active work on this topic continues. A potential speed-up can be achieved either by eliminating the fine alignment step in the right circumstances or by skipping the pre-alignment if the scans are separated by a very short time. As an added benefit of processing the {LiDAR} data on board, obstacle avoidance can be performed on them at the same time with a reduced performance hit.

\subsection{Further work}

Further work has to be invested into merging the two localisation sources into one, coherent position estimation. The candidates outlined in the introductory section of this articles are considered: Kalman and particle filters. Finding a parameter more robust than HDOP which could be used to tune the filters online is a desirable result.

\section*{ACKNOWLEDGMENT}

The authors wish to acknowledge the kind help and advice received during the trials from the MIT Sailing Pavilion team and the MIT Sea Grant colleagues.

\bibliographystyle{IEEEtrans}

\bibliography{root}

\end{document}